\documentclass{article}
\usepackage[preprint, nonatbib]{neurips_2020}
\pdfoutput=1
\usepackage[utf8]{inputenc} 
\usepackage[T1]{fontenc}    
\usepackage{hyperref}       
\hypersetup{pdfauthor={Sharma, Challa, Gupta, Murty},pdftitle={Higher-Order Relations Skew Link Prediction}}

\usepackage{wrapfig}

\usepackage{url}            
\usepackage{booktabs}       
\usepackage{amsfonts}       
\usepackage{nicefrac}       
\usepackage{microtype}      

\usepackage{amsmath,amsfonts}
\usepackage{amsthm}
\usepackage{amsmath, amssymb}
\usepackage{graphicx}
\usepackage{enumerate}
\usepackage{xcolor}
\usepackage{multirow}
\usepackage[linesnumbered, ruled, vlined]{algorithm2e}
\usepackage{subfig}

\newtheorem{theorem}{Theorem}
\newtheorem{proposition}{Proposition}

\newcommand{\powerset}{\mathcal{P}}
\newcommand{\dataee}{\texttt{email-Enron}}
\newcommand{\datachs}{\texttt{contact-high-school}}
\newcommand{\datacps}{\texttt{contact-primary-school}}
\newcommand{\datandc}{\texttt{NDC-substances}}
\newcommand{\datatms}{\texttt{tags-math-sx}}

\newcommand{\algoPA}{\textit{PA}}
\newcommand{\algoAA}{\textit{AA}}
\newcommand{\algoCN}{\textit{CN}}
\newcommand{\algoJC}{\textit{JC}}
\newcommand{\algoRA}{\textit{RA}}
\newcommand{\algoSR}{\textit{SR}}

\SetKwInput{KwInput}{Input}
\SetKwInput{KwOutput}{Output}

\title{Higher-Order Relations Skew Link Prediction in Graphs}

\author{%
  Govind Sharma \qquad Aditya Challa \qquad Paarth Gupta \qquad M. Narasimha Murty\\~\\
  \texttt{\href{mailto:govinds@iisc.ac.in}{govinds@iisc.ac.in}, \href{mailto:adityachalla20@gmail.com}{adityachalla20@gmail.com}, \{\href{mailto:paarthgupta@iisc.ac.in}{paarthgupta}, \href{mailto:mnm@iisc.ac.in}{mnm}\}@iisc.ac.in}\\~\\
  Department of Computer Science and Automation\\
  Indian Institute of Science, Bangalore\\
  Karnataka 560012, India
}

\begin{document}

\maketitle
\begin{abstract}
    The problem of link prediction is of active interest.
    The main approach to solving the link prediction problem is based on heuristics such as Common Neighbors (CN) -- more number of common neighbors of a pair of nodes implies a higher chance of them getting linked.
    In this article, we investigate this problem in the presence of higher-order relations.
    Surprisingly, it is found that CN works very well, and even better in the presence of higher-order relations.
    However, as we prove in the current work, this is due to the CN-heuristic overestimating its prediction abilities in the presence of higher-order relations.
    This statement is proved by considering a theoretical model for higher-order relations and by showing that AUC scores of CN are higher than can be achieved from the model.
    Theoretical justification in simple cases is also provided.
    Further, we extend our observations to other similar link prediction algorithms such as Adamic Adar.
    Finally, these insights are used to propose an adjustment factor by taking into conscience that a random graph would only have a best AUC score of 0.5.
    This adjustment factor allows for a better estimation of generalization scores.
\end{abstract}

\section{Introduction}
\label{sec:intro}

The problem of link prediction (LP) is described as follows: ``{Given a set of objects $V$, and a set $E \subseteq  \powerset_2(V)$\footnote{$\powerset_2(V)$ denotes all 2-subsets of $V$.} of (partial) links among them, predict new/missing links among $V$}''.
This is naturally modelled as a simple graph $G = (V, E)$.
Ever since the seminal work on this problem \cite{Liben-Nowell2003}, it has seen constant advancements \cite{Wang2015, Martinez2016}.
Standard LP algorithms are based on heuristics such as Common Neighbors (CN)~\cite{Newman2001}, which posits that more number of common neighbors imply a higher chance of link between a pair of nodes, or Adamic Adar (AA)~\cite{Adamic2003}, a normalized version of the CN approach.
These heuristics are known to work dramatically well for simple datasets \cite{liben2003link, sarkar2011theoretical, cohen2015axiomatic}.
In this article, we consider the LP problem (and algorithms for the same) in the presence of \emph{higher order relations.}

Higher-order relations are modelled using a structure called a \emph{hypergraph}~\cite{Berge1984}, which is defined as a tuple of a vertex set, $V$ and a collection  $F \subseteq 2^V$ of its subsets, \textit{viz.}, \textit{hyperedges}.
Essentially, it extends the traditional notion of usual graphs by allowing edges of a higher order (\textit{i.e.}, those containing arbitrary numbers of nodes).
These structures are typically used to model higher-order real world relations.
It is also a common practice to reduce a hypergraph to simple graphs by considering all possible $2$-subsets of hyperedges to get edges $E:= \cup_{f \in F} \mathcal{P}_2(f)$.
This procedure is referred to as \emph{clique expansion}~\cite{Agarwal2006}.
If $H$ denotes a hypergraph, we denote its corresponding clique-expanded graph as $\eta(H)$.

\begin{figure}
  \centering
  \begin{minipage}{0.56\linewidth}
    \subfloat[]{\centerline{\includegraphics[width=0.8\linewidth, trim=1cm 0 5cm 0, clip]{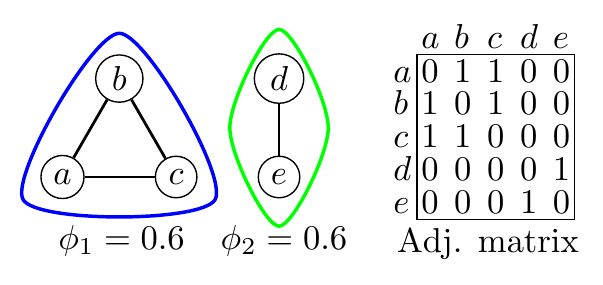}}}
\end{minipage}%
  \begin{minipage}{0.44\linewidth}
    \subfloat[]{\centerline{{\includegraphics[width=0.8\linewidth, trim=7.3cm 0 0 0, clip]{img_new/figure1V2a.pdf}}}}
\end{minipage}

  \caption{Figure illustrating the effect of higher-order relations.
  (a) A toy hypergraph where set of vertices is $\{a,b,c,d,e\}$.
  $\{a,b,c\}$ denotes one hyperedge (in blue) which appears with probability $\phi_1=0.6$ and $\{d,e\}$ is another one (in green) appearing with the same probability $\phi_2=0.6$.
  The four arcs $a\sim b$, $b\sim c$, $a\sim c$, and $d\sim e$ (in black) denote edges of the graph formed by the hypergraph's clique expansion~\cite{Agarwal2006}, whose adjacency matrix has also been shown.
  (b) CN scores are calculated using the \emph{leave-one-out} method and are shown as a matrix.
  Here, CN predicts that the link $b \sim c$ ($CN=1$) is more probable than $d \sim e$ ($CN=0$).
  However, this is \emph{not} the case as both these links occur with the same probability (\textit{viz.}, $0.6$).
  An evaluation of the CN principle, however, does not take this into account, and when its ROC curve is plotted \textit{w.r.t.} the adjacency matrix, a predictive $AUC$ (area of the shaded region) of $0.875$ is estimated.
  While empirically, one only obtains an $AUC$ of $0.5$.
  Hence it is clear that CN overestimates its ability to predict links.
  This is a simple illustration to show that higher-order relations skew LP algorithms.
  In this article, we formalize these notions and provide a novel evaluation method correcting this effect.}
  \label{fig:1}
\end{figure}

We show that in the presence of higher-order relations, LP algorithms do not generalize well.
Moreover, we prove that evaluation of LP algorithms in the presence of higher-order relations overestimates their prediction capability.
As a simple example, consider the network in Figure~\ref{fig:1}.
Let it consist of $5$ vertices named $a$--$e$.
Also assume that it consists of two hyperedges $\{a,b,c\}$ and $\{d,e\}$, appearing with probabilities $\phi_1=0.6$ and $\phi_2=0.6$ respectively.
This has been depicted in Figure \ref{fig:1} using the blue and green enclosures respectively.
Let us apply CN over this example by predicting links using the \emph{leave-one-out} method of evaluation and subsequently compute its $AUC$ score.
CN asserts that link $b\sim c$ is more probable than link $d\sim e$ since the former has more common neighbors than the latter.
However this is \emph{not} the case since both these links occur with probability $0.6$.
Moreover, the evaluation of CN for this example \emph{does not} take this into account, thereby estimating the predictive $AUC$ score to be $0.875$.
While actually, one can only obtain an $AUC$ of $0.5$ -- a fact that can be verified empirically.
Thus CN overestimates its own predictive capabilities. In this article we formalize these notions and provide both theoretical and empirical support to these observations. Moreover, we also provide a novel evaluation method, proposing an \textit{adjustment-factor} to correct the predictive scores.

In Section \ref{sec:2}, we propose a simple mathematical model for higher-order relations, which is used for analysis in the rest of the article.
For completeness, we compare this to existing latent space models for link prediction as well.
Then in Section \ref{sec:3}, we use the model in Section \ref{sec:2} to prove empirically that the LP heuristics CN and AA overestimate their generalization-ability.
This is justified theoretically considering simple cases.
In Section \ref{sec:4}, we propose a new evaluation scheme which takes into account the higher-order relations. Finally in Section \ref{sec:5} we discuss the implications of this work and discuss future directions.
The main contributions of this article can be summarized as follows:
\begin{enumerate}
    \item We prove that higher-order relations skew link prediction.
    In particular, we show that standard heuristics such as CN and AA do not generalize well in the presence of higher-order relations.
    Moreover, we show that the evaluation of these methods also do not take this into consideration, thereby overestimating their ability to predict links.
    \item To provide better estimates of the generalization performance, we propose a novel approach to compute an \emph{adjustment factor} to correct the generalization scores.
\end{enumerate}   

\section{A Mathematical Model for Higher-Order Relations}
\label{sec:2}
In this section, we provide a simple model for modelling higher-order relations, which is used in the rest of the article for analysis and simulation.
This has been adapted from Turnbull et al.~\cite{turnbull2019latent}, the main difference being our assumption that the latent space is fixed.
Recall that to specify a hypergraph, one needs to specify the set of objects $V$ and the subsets of $V$ chosen to be hyperedges, $F \subseteq 2^{V}$.

As discussed earlier, let $V = \{1,2,\cdots,n\}$ denote a set of objects.
For each element $i \in V$, assume there exists an underlying vector $u_i$ in the \emph{latent space} $\mathbb{R}^{d}$.
To model the hyperedges in this space, we assume that their sizes/cardinalities (number of objects in a hyperedge) lie in the set $\{1,2,\cdots,k\}$, where $1 \leq k \leq n$.
Let $r_1, r_2, \cdots, r_k \in \mathbb{R}$ be called as the \textit{radii} corresponding to hyperedge sizes $1, 2, \cdots, k$ respectively.
As per our model, we define a subset $f \in 2^V$ to be a ``potential hyperedge'' if and only if there exists a ball of radius $r_{\lvert f \rvert}$ encompassing the set of latent vectors of its containing objects.
That is for a given subset $f \in 2^V$, we have:
\begin{equation}
  f \in \overline{F} \iff \exists x\in \mathbb{R}^d \text{ s.t. } \{u_i\}_{i\in f} \subseteq B\left(x, r_{|f|}\right),
\end{equation}
where $\overline{F}$ denotes the set of \emph{potential hyperedges}.
It is unlikely that all the potential hyperedges belong to the final hypergraph.
Hence, we introduce probabilities $\phi_1, \phi_2, \cdots, \phi_k \in [0, 1]$ corresponding to hyperedge sizes $1, 2, \cdots, k$ respectively.
The final set of hyperedges would then be given by:
\begin{equation}
  f \in F \iff f \in \overline{F} \text{ and } \text{Bernoulli}(\phi_{\lvert f \rvert}) = 1,
\end{equation}
that is, a potential hyperedge would be in the final set of hyperedges with probability $\phi_{\lvert f \rvert}$.

\begin{figure}[t]
    \begin{minipage}{0.333\linewidth}
        \subfloat[$U = \{u_i\}_{i=0}^{9}$]{\centerline{\centerline{\includegraphics[trim={6cm 11cm 6cm 11cm}, clip, width=0.8\linewidth]{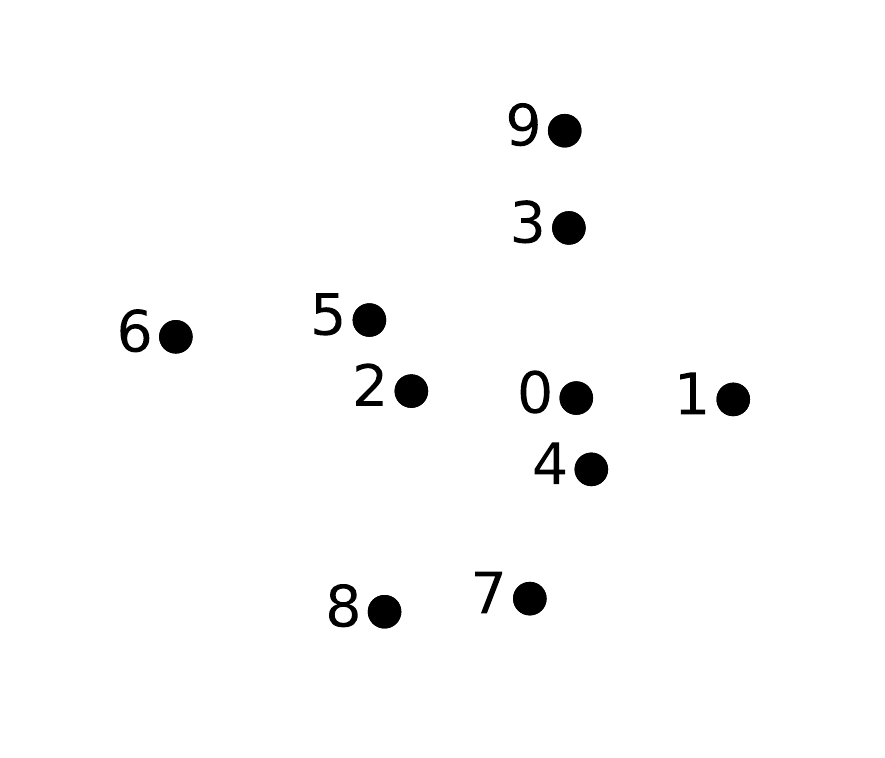}}}}
    \end{minipage}%
    \begin{minipage}{0.333\linewidth}
        \subfloat[$\overline{F}(U, r)$]{\centerline{\includegraphics[trim={6cm 11cm 6cm 11cm},clip,width=0.8\linewidth]{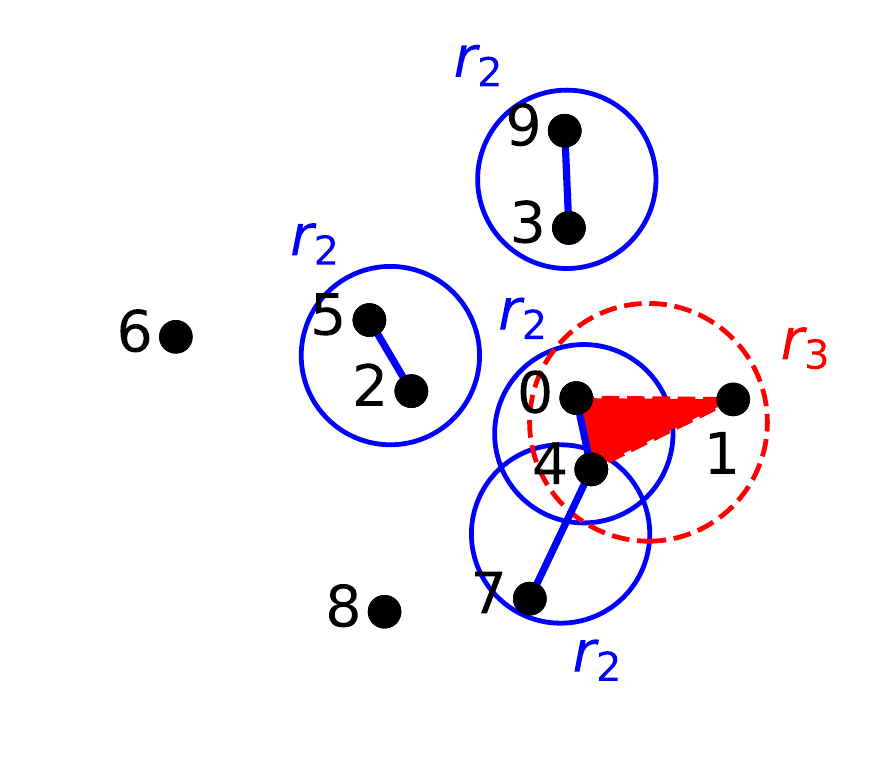}}}
    \end{minipage}%
    \begin{minipage}{0.333\linewidth}
        \subfloat[$F(U, r, \Phi)$]{\centerline{\includegraphics[trim={6cm 11cm 6cm 11cm},clip,width=0.8\linewidth]{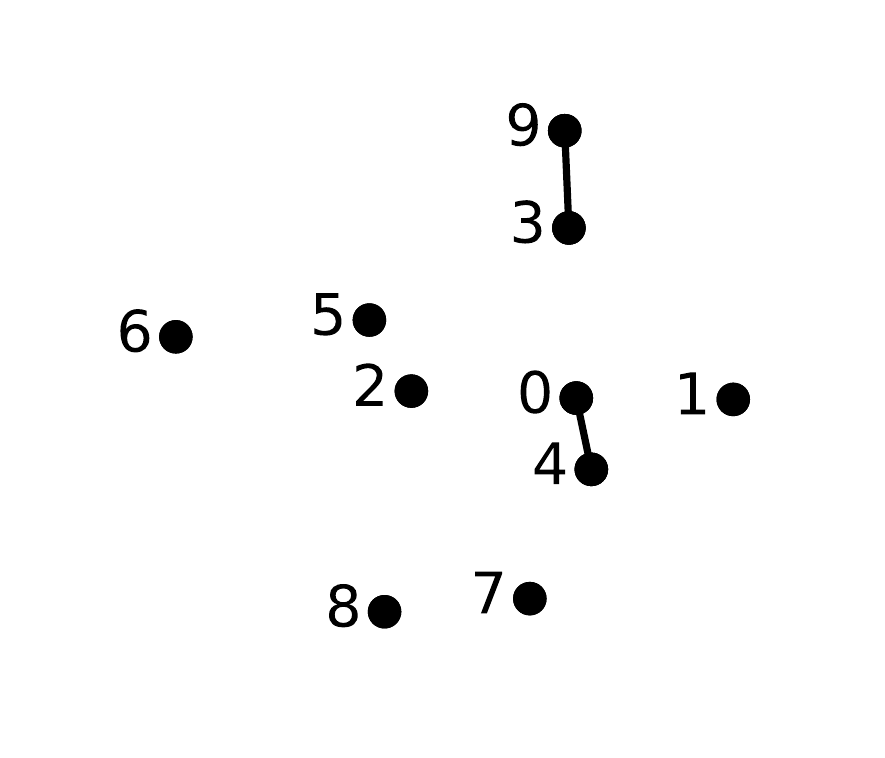}}}
    \end{minipage}%
    \caption{An illustration of the \textit{hypergraph latent space model} described in Section~\ref{sec:2} for ten vertices $1$--$10$. (a) The 2-dimensional representation $U$ of \textit{vertices}. (b) Radii $r=(r_2, r_3)$ are picked and \textit{potential hyperedges} $\overline{F}$ are generated: four $2$-sized hyperedges $\{0, 4\}$, $\{2, 5\}$, $\{3, 9\}$, $\{4, 7\}$ \textit{w.r.t.} radius $r_2$, and one $3$-sized hyperedge $\{0, 1, 4\}$ \textit{w.r.t.} radius $r_3$. Note the balls of radii $r_2$ (blue solid) and $r_3$ (red dotted) encompass vertices that form the hyperedges. (c) Finally, \textit{hyperedges} $F$ are sampled from $\overline{F}$ via distribution $\Phi$; in this case, we have $F = \{\{0, 4\}, \{3, 9\}\}$.}
    \label{fig:ls_hyg}
\end{figure}

To generate a hypergraph using the model above, one can start with an arbitrary representation $U$, say from a normal distribution with mean $\mathbf{0}_d$ (zero vector of size $d$) and co-variance $\mathbf{I}_d$ (identity matrix of size $d \times d$), and pick $r$ to be the fixed percentiles of all the pairwise distances. The hyperedges are generated using:
\begin{enumerate}
  \item Set $s = 2$, and $\overline{F} = \{\}$.
  \item Start with radius $r_s$, and select all groups with distance $\leq 2 r_s$ as $s$-sized hyperedges and add them to $\overline F$.
  \item Obtain the cliques of size $s$ in this hypergraph.
  \item Repeat steps 2 and 3 above for radii of a higher order ($s > 2$) to obtain the set of potential hyperedges for all sizes $2, 3, \hdots, k$, and ultimately get $\overline{F}$.
  \item Finally, select each hyperedge $f \in \overline{F}$ with a probability $\phi_{\lvert f \rvert}$ to get $F$.
\end{enumerate}
Refer to Figure~\ref{fig:ls_hyg} for an illustration of the foregoing procedure.
This procedure generates a \emph{Vietoris–Rips Complex}~\cite{reitberger2002leopold}, which is equivalent to a \v{C}ech complex~\cite{ghrist2014elementary,edelsbrunner2010computational,turnbull2019latent}. Moreover, there exist faster algorithms to achieve this as well~\cite{zomorodian2010fast}.

\subsection{Relation to Hoff's Latent Space Model}

\begin{figure}[t]
  \begin{minipage}{0.5\linewidth}
    \subfloat[]{\centerline{\includegraphics[width=0.95\linewidth, trim={ 0 7cm 0 7cm}, clip]{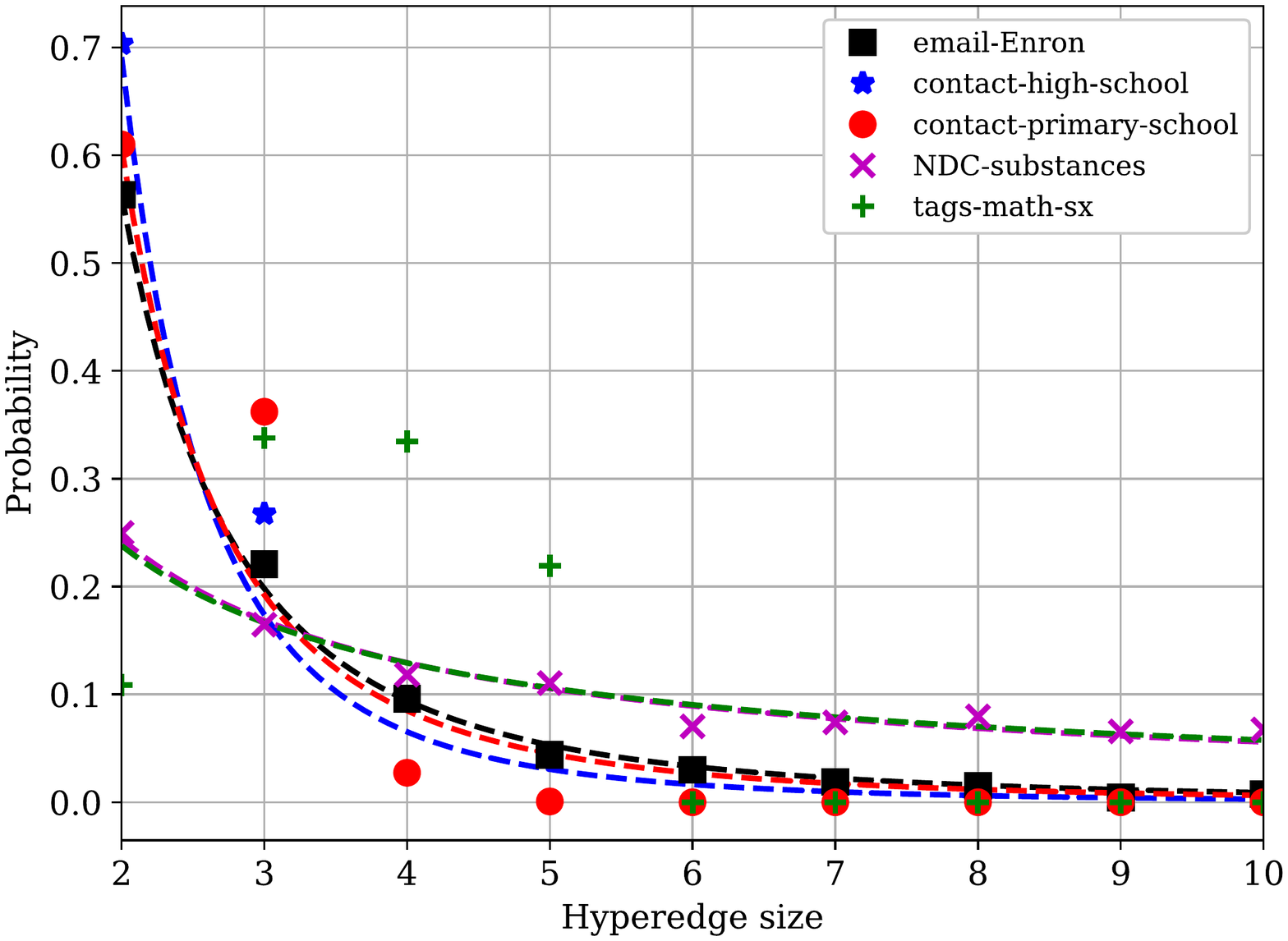}}}
\end{minipage}%
  \begin{minipage}{0.5\linewidth}
    \subfloat[]{\centerline{{\includegraphics[trim={0.2cm 0 1.4cm 1.2cm}, clip, width=\linewidth]{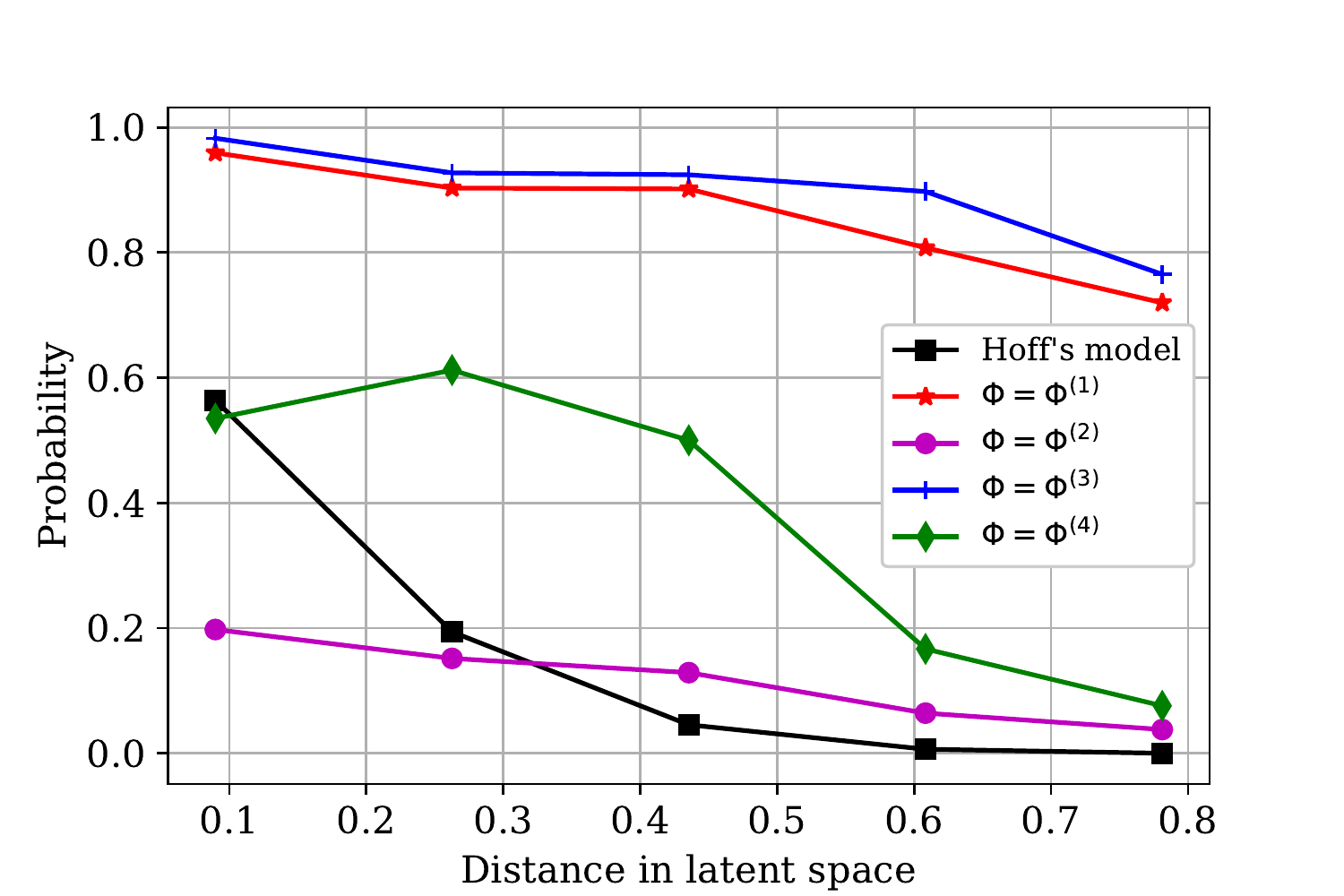}}}}
\end{minipage}%
  \caption{(a) Hyperedge size ($|f| = k$) distribution in a few real-world datasets. Each distribution is fitted with a power-law $1/(k^\zeta)$ truncated between $2$ and $10$, with best-fit $\zeta$-values being as follows: \textit{email-Enron}: $2.58$, \textit{contact-high-school}: $3.43$, \textit{contact-primary-school}: $2.83$, \textit{NDC-substances}: $0.91$. More details about the datasets are available in Table~\ref{tab:relocated_aucs}. (b) Edge-generation probabilities plotted against distance between (the latent space representations of) incident nodes thereof. We consider different choices for $\phi_k$ and look at the distribution of the edges: $\phi^{(1)}_k = 1/k^2$, $\phi^{(2)}_k = 1/(1 + e^{\alpha(r_k - \gamma)})$, $\phi^{(3)}_k = 0.1$, $\phi^{(4)}_k$ as per $\overline{F}$. The latent space $U$ is generated using a 2-dimensional normal distribution with mean $\mathbf{0}_2$ and covariance $\mathbf{I}_2$ (identity matrix of size $2 \times 2$). The radii are picked to be the $1$-, $5$-, $9$-, and $13$-percentiles of all the distances between the points. Observe that for large distances (ranging between $0.4$--$0.6$), Hoff's model underestimates the number of edges. }
  \label{fig:hyg_sizes}
\end{figure}

Classically, link prediction in simple graphs has been modelled using what we call the \emph{Hoff's model}, which is described in Hoff et al.~\cite{doi:10.1198/Hoff2002Latent}. Authors in Sarkar et al.~\cite{sarkar2011theoretical} use this model to provide theoretical justifications to three LP heuristics.
However, as we shall shortly see, Hoff's model underestimates the higher-order relations.
Basically, it assumes that two vertices $i$ and $j$ are linked to each other with probability
\begin{equation}
  P(i \sim j) = \frac{1}{1 + \exp(\alpha(\| u_i - u_j \| - \gamma))},
\end{equation}
where $\alpha$ and $\gamma$ are the model's parameters and $u_i$ and $u_j$, the vertices' latent vectors.
Thus, a hyperedge $f \in F$ would show its existence via Hoff's model if all $2$-subsets of $f$ (\textit{i.e.}, edges comprising the clique over nodes in $f$) get selected by the model.
We have, 
\begin{equation}
      P(i \sim j, \forall i, j \in f, i \neq j) = \prod_{\substack{i,j \in f\\i\neq j}} P(i \sim j) 
      = \prod_{\substack{i,j \in f\\i\neq j}} \frac{1}{1 + \exp(\alpha(d_{ij} - \gamma))},
\end{equation}
where $d_{ij} := \|u_i - u_j\|$.
Observe that if $|f| = k$, then this has $k(k-1)/2$ factors in the product and thus reduces as $\mathcal{O}(1/C^{k(k-1)/2})$ for some constant $C > 1$.
Thus, the number of hyperedges reduces exponentially \textit{w.r.t.} hyperedge size $k$ according to Hoff's model.
However, in most real-world hypergraphs, the number of hyperedges have been observed to follow a power law, $ 1/k^{\zeta}$ as shown in Figure~\ref{fig:hyg_sizes}(a), where $\zeta > 0$ varies from domain to domain.
Hence, we know that Hoff's model does not capture higher-order relations well.
Another implication of this observation is that Hoff's model also underestimates number of long distance edges.
To illustrate this, we compare the probabilities of generating an edge of distance $d$ by both the models. We show this for four choices of $\Phi$ in Figure \ref{fig:hyg_sizes}(b).

{\noindent
\textbf{Remark:} The choices of $\Phi$ are dictated by conventional wisdom. (i) $\phi^{(1)}_k := 1/k^2$ is used since in real datasets, a power law size distribution is observed. (ii) $\phi^{(2)}_k := 1/(1 + \exp(\alpha(r_k - \gamma)))$ is used since this is the probability that an edge with distance $r_k$ is picked. (iii) For completeness, we also consider $\phi^{(3)}_k = 0.1$. (iv)  Another option is to take  $\phi^{(4)}_k$ as per the distribution in $\overline F$.
}

\section{Effect on the Evaluation of Link Prediction}
\label{sec:3}
In this section we shall use the model from Section~\ref{sec:2} to analyze the performance of the LP heuristics - Common Neighbors (CN) and Adamic Adar (AA). Specifically, we show that \emph{these heuristics overestimate their ability to predict links}.

\subsection{Capturing the Generalization Error}
Recall that the hypergraph model uses a triplet $(U, r, \Phi)$ of vertex representation vectors $U$, size-specific radii $r$, and hyperedge selection probability distribution $\Phi$ to obtain the hypergraph $H = (V, F)$.
This is converted to a simple graph $\eta(H)$ using \emph{clique expansion}~\cite{Agarwal2006}.
We then have the following proposition:
\begin{proposition}
  \label{prop:1}
  In the model described in Section~\ref{sec:2}, we have
  \begin{equation}
      P(i \not\sim j) = \Pi_{k} (1 - \phi_k)^{S_k(i,j)},
  \end{equation}
  where $S_k(i,j)$ is the number of $k$-sized hyperedges in $\overline{F}$ which contain both vertices $i$ and $j$ given $U$ and $r$.
  Clearly, we also have 
  \begin{equation}
  \label{}
      P(i \sim j) = 1 -  \Pi_k (1 - \phi_k)^{S_k(i,j)}.
  \end{equation}
\end{proposition}
\begin{proof}
Please refer to Appendix~\ref{app:proof:prop}.
\end{proof}

\begin{figure}[!t]
  \begin{minipage}{0.5\linewidth}
      \subfloat[CN]{\centerline{{{\includegraphics[width=0.8\linewidth, trim={ 0 4.5cm 0 4.5cm}, clip]{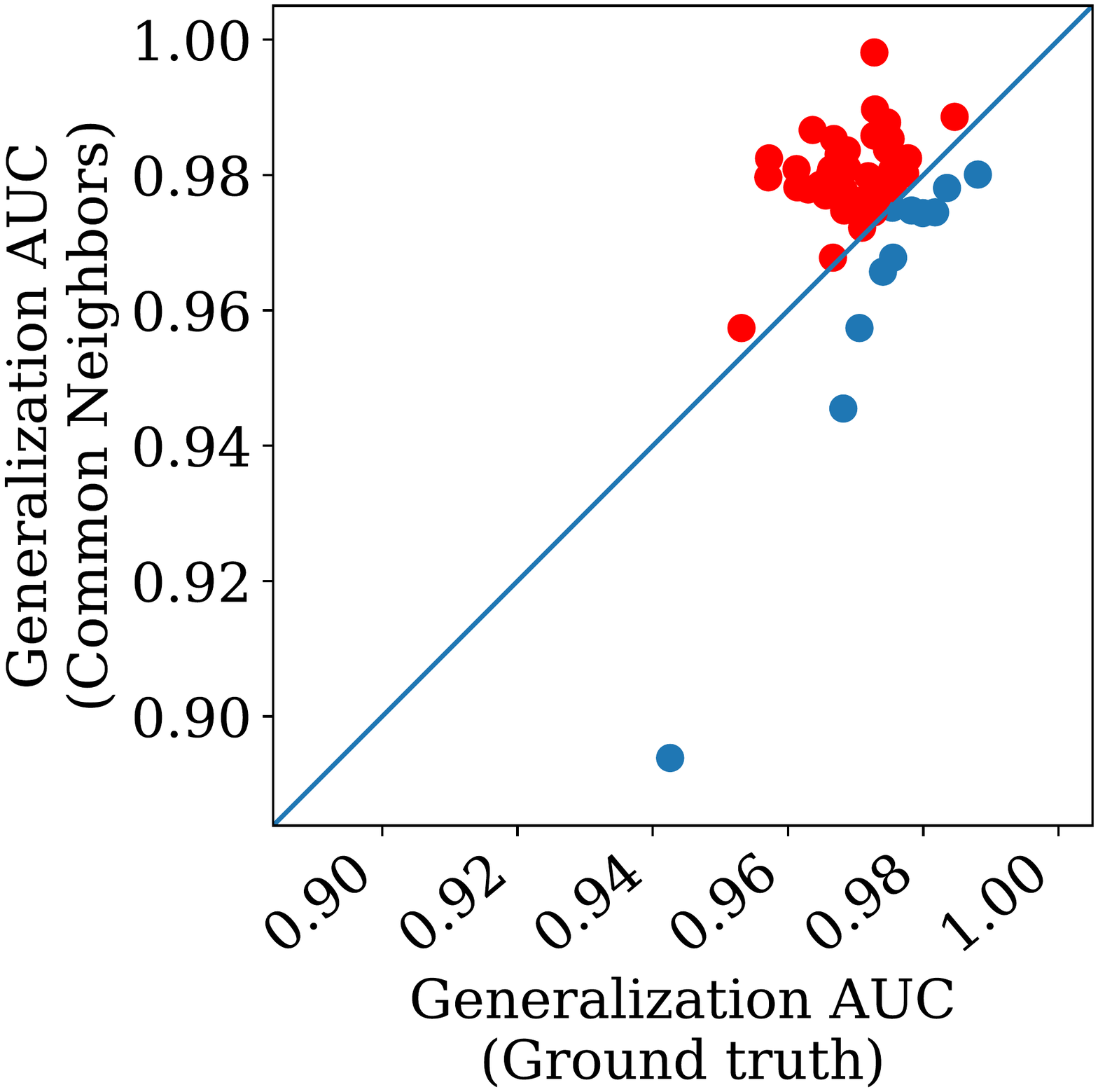}}}}}
  \end{minipage}%
  \begin{minipage}{0.5\linewidth}
      \subfloat[AA]{\centerline{{\includegraphics[width=0.8\linewidth, trim={ 0 4.5cm 0 4.5cm}, clip]{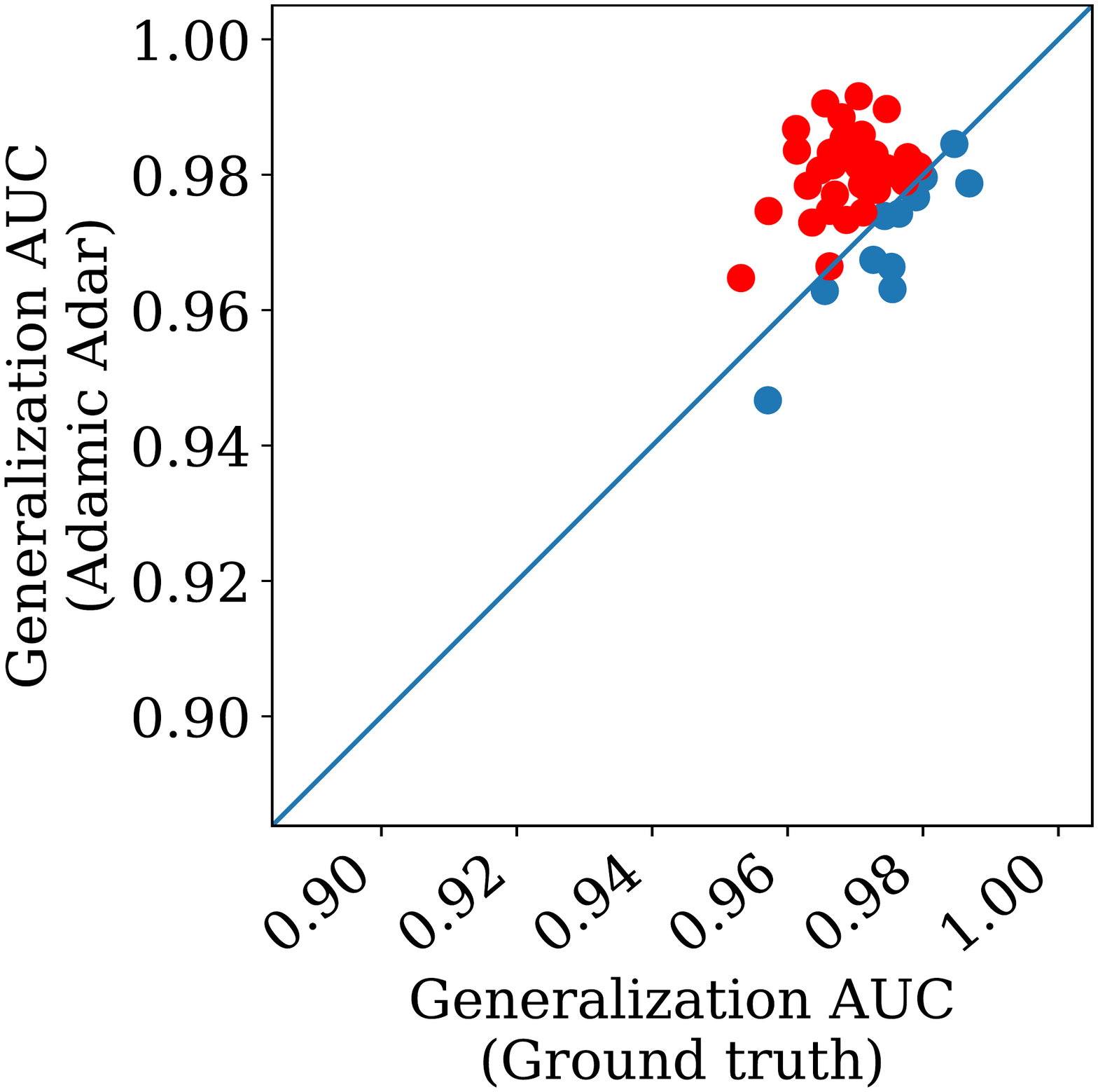}}}}
  \end{minipage}%
  \caption{Generalization performances of heuristic LP methods. Observe that in several cases (the ones marked in red), the generalization performance predicted by both CN and AA is higher than the actual one derived from the model. This shows that in presence of hyperedges, existing methods overestimate the generalization performance. (a) Comparison with Common Neighbors (b) Comparison with Adamic Adar. }
  \label{fig:2}
\end{figure}

Using Proposition~\ref{prop:1} one can compute the probability of a link between two vertices $i$ and $j$.

\subsection{The Behavior of Link Prediction Heuristics}
On the other hand, the LP heuristics CN and AA dictate that this probability is proportional to the number of common neighbors.
Figure \ref{fig:2} shows the scatter plots between the AUC scores obtained from the model and those by the LP heuristics.
Observe that in several cases (marked red) the generalization performance as estimated by the LP heuristics is \emph{higher} than the ground truth probabilities.
However, theoretically, \emph{the generalization performance of any algorithm cannot be better than the one obtained using the estimates in eq.~\eqref{eq:prob-generalization}}.
This shows empirically that LP heuristics such as CN/AA overestimate their ability to predict links.

\begin{wrapfigure}{r}{0.5\textwidth}
  \begin{minipage}[b]{0.5\linewidth}
      \subfloat[]{\centerline{\includegraphics[trim={1cm 0 0 0},clip, width=0.55\linewidth]{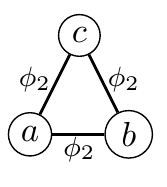}}}
  \end{minipage}%
  \begin{minipage}[b]{0.5\linewidth}
      \subfloat[]{\centerline{\includegraphics[trim={1.44cm 0.1cm 0.5cm 0.5cm}, clip, width=0.6\linewidth]{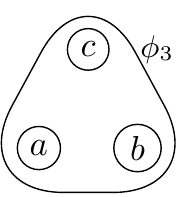}}}
  \end{minipage}%
  \caption{Example hypergraphs. Both the examples consist of 3 vertices $a$, $b$, $c$. Assume that the set of potential hyperedges is $\overline{F} = \{\{a,b\}, \{a,c\}, \{b,c\}, \{a,b,c\}\}$, and hyperedge-selection probabilities are $\Phi = (\phi_2, \phi_3)$. (a) The hypergraph where the three 2-edges are possible with probability $\phi_2 > 0$ (and $\phi_3$ is fixed to zero). (b) The hypergraph where the 3-edge is possible with probability $\phi_3 > 0$ (and $\phi_2$ is set to zero). We have that the AUC scores of the model for CN match in (a), which is equal to $0.5$. However in (b), we have that the AUC score of CN is $1$, while the AUC score from the model is just $0.5$. Thus, CN overestimates its ability to predict the links in presence of higher-order relations.}
  \label{fig:4}
\end{wrapfigure}

To understand better why this is the case, consider two scenarios: (i) A simple graph without higher-order relations as shown in Figure~\ref{fig:4}(a), and (ii) A hypergraph with a single hyperedge of size $3$ as shown in Figure~\ref{fig:4}(b). We now closely analyze the AUC scores of the CN heuristic. 
For this, we require the following notation: Let $Z_{\sim} := (CN/(i\sim j))$ denote the random variable which counts the number of common neighbors between $i$ and $j$ when it is known that $i \sim j$ ($i$ is linked to $j$). Similarly, let $Z_{\not\sim} := (CN/(i\not\sim j))$ be a random variable counting common neighbors when $i \not\sim j$.
Using this notation and a result from Fawcett et al.~\cite{fawcett2006introduction}, we have that the AUC score can be computed using:
\begin{equation}
  \label{eq:auc charac}
  AUC = P(Z_{\sim} > Z_{\not\sim}/ (Z_{\sim} \neq Z_{\not\sim})).
\end{equation}

Now, in the case of Figure \ref{fig:4}(a), which is essentially a graph with no higher-order relations, we have, 
\begin{equation}
  Z_{\sim} = Z_{\not\sim} = \begin{cases}
      1 & \text{\textit{w. p. }} \phi_{2}^{2}, \\
      0 & \text{\textit{w. p. }} 1 - \phi_{2}^{2} \\
  \end{cases}
\end{equation} 
So, it is easy to see that we have $AUC$ score to be $0.5$ as can be computed from eq. \eqref{eq:auc charac}. In other words, the CN heuristic estimates that it cannot predict well whether node $a$ would be linked to node $b$.
This prediction matches with the ground-truth, since in the first place, the link between node $a$ and node $b$ is randomly present with probability $\phi_2 > 0$. Now, consider the scenario when a higher-order relation is present in the network, \textit{i.e.}, it is a hypergraph, as shown in Figure \ref{fig:4}(b).
In this case, the link between nodes $a$ and $b$ appears randomly with probability $\phi_3$, and hence, any heuristic should not be able to predict the link with an AUC score of more than $0.5$. 
However, in this case observe that: 
\begin{equation}
  Z_{\sim} = \begin{cases}
      1 & \text{\textit{w. p. }} 1,\\
      0 & \text{\textit{w. p. }} 0 \\
  \end{cases} \text{ ~~and } 
  Z_{\not\sim} = \begin{cases}
      1 & \text{\textit{w. p. }} 0, \\
      0 & \text{\textit{w. p. }} 1. \\
  \end{cases}
\end{equation}
And from eq. \eqref{eq:auc charac}, we have that $AUC$ score is $1$.
In other words, the CN heuristic estimates it can predict perfectly whether the link between nodes $a$ and $b$ exists or not.
But, it is known that this is not possible since the existence of a link between $a$ and $b$ is, by construction, random with probability $\phi_3 > 0$.
Hence, this justifies the empirical observation that LP heuristics CN and AA overestimate their ability to predict links in presence of higher-order relations (hyperedges). 
We state and prove a similar argument for a generic hypergraph in Theorem~\ref{thm:1}, making our case even stronger.

\begin{theorem}
\label{thm:1}
Let $(U,r,\phi)$ denote the hypergraph model, where $\phi_2 = 0$ and $\phi_i > 0$ for all $i > 3$. Then the AUC score of CN is strictly greater than 0.5. 
\end{theorem}

Note that since any link $i\sim j$ can occur only with probability $\phi_{\lvert h \rvert}$ where $h \supset \{i,j\}$, the best possible score can be only $0.5$.  

\begin{proof}
Please refer to Appendix~\ref{app:proof:1}.
\end{proof}

\section{Better Evaluation of Link Prediction Methods}
\label{sec:4}
In the previous section, we have proved that higher-order relations skew scores provided by LP heuristics.
More particularly, we saw that they tend to overestimate their capability of generalization.
In this section, we provide a method to better estimate this generalization-ability. 

\begin{theorem}
\label{thm:2}
On an Erdos-Renyi graph, the best AUC a link prediction method can achieve is of $0.5$.
\end{theorem}
\begin{proof}
Please refer to Appendix~\ref{app:proof:2}.
\end{proof}

The main idea is as follows, that relies on this particular premise: ``\textit{On a random version of a given graph, the best AUC a link prediction method can achieve is of $0.5$}''.
Thus, given a hypergraph $H$, one can construct a \textit{randomized version} $H_{rand}$ of $H$, and expect the link prediction AUC on its clique-expanded graph $\eta(H_{rand})$ to be around $0.5$.
Now, as noted in the previous section, we know that a typical LP algorithm gives a higher-than-expected AUC score on any graph expanded from a hypergraph.
Thus, we compute an \emph{adjustment factor} $AF$, which we define as the ratio of AUC score $AUC(H_{rand})$\footnote{For a hypergraph $X$, $AUC(X)$ denotes the AUC score obtained on its clique-expanded graph $\eta(X)$.} obtained on the randomized hypergraph and the ideally expected score \textit{viz.}, $0.5$ on it.
Finally, this adjustment factor is used to compute an \emph{adjusted AUC score} $AUC_{adj}(H)$ on the original hypergraph $H$.

To achieve this, we first make multiple runs of a \textit{hyperedge relocation algorithm} (Algorithm~\ref{alg:relocated_hyg}) on a given hypergraph $H$ to obtain multiple relocated versions $H_{rel}$ of the same.
Basically, for each hyperedge in the original hypergraph $H$, we add a same-sized random hyperedge to $H_{rel}$.
This ensures that the core statistics of the network remains the same.
However, since the hyperedges are added randomly, any LP algorithm should only have achieved a score of $0.5$.
The \textit{adjustment factor} and accordingly, an \emph{adjusted AUC score} can then be computed using relocated AUC $AUC_{rel} := AUC(H_{rel})$ as:
\begin{equation}
\label{eq:adj_factor}
  AF(H) = \frac{AUC_{rel}}{0.5}\hspace{2cm} AUC_{adj}(H) = \frac{AUC(H)}{AF(H)}
\end{equation}

\begin{algorithm}[H]
  \caption{Generates randomized version of a hypergraph $H$, $H_{rel}$. This is referred to as the \emph{relocation algorithm}. Every hyperedge in $H$ is relocated to a randomly selected new one, thereby constructing a \emph{relocated hypergraph} $H_{rel}$. Multiple runs of this algorithm on the same network is used to estimate baselines for an LP algorithm.}
  \label{alg:relocated_hyg}
        \DontPrintSemicolon
  \KwInput{\text{Original hypergraph,} $H = (V, F)$\newline
  {\text{LP algorithm,} $X$}}
  \KwOutput{$H_{rel}$}
      \BlankLine
      $F_{rel} \gets \{\}$\;
      \For{$f \in F$}{
          $f_{rel} \gets$ \textsc{RandomSubset}$(V, |f|)$\;
          $F_{rel} \gets F_{rel} \cup \{f_{rel}\}$\;
      }
      $H_{rel} \gets (V, F_{rel})$\;
  \Return{$H_{rel}$}
\end{algorithm}


\begin{table*}
  \caption{Popular neighborhood-based link prediction (LP) algorithms' AUC scores (\%) for five real-world hypergraphs ($H$) and their relocated versions ($H_{rel}$), providing new baselines (\textit{i.e.}, relocated AUCs $AUC_{rel}$) and corrected performance scores (adjusted AUCs, $AUC_{adj}$) for them.
  Observe that the newer baselines are different across different datasets per method.
  Using the mean relocated AUC scores, we compute adjustment factors $AF$ and ultimately report an adjusted AUC score $AUC_{adj}$ as per eq.~\eqref{eq:adj_factor}.
  Also shown are number of nodes $|V|$ and hyperedges $|F|$, whose size distributions are depicted in Figure~\ref{fig:hyg_sizes}.
  All datasets have been taken from Benson et al.~\cite{Benson2018}. Adjustment factors that are lowest for a single dataset are \textbf{bold-faced}.
  In addition, four algorithm-pairs have been marked using superscripts $a$, $b$, $c$, and $d$; these show performance-reversal \textit{w.r.t.} $AUC$ and $AUC_{rel}$.}
  \label{tab:relocated_aucs}
  \centering
  \footnotesize
  \begin{tabular}{llcccc}
\toprule
&LP&Orig. AUC & Relocated AUC &  &Adj. AUC\\
\textbf{Dataset $H$} &\hspace{-0.5cm}algorithm& \textbf{$AUC$} & \textbf{$AUC_{rel}$} &  $AF$ & $AUC_{adj}$\\
\midrule
\midrule
\multirow{6}{*}{\shortstack[l]{\dataee\\~\\$|V|=148$\\$|F|=1{,}436$}} & \algoPA{} &  $82.46$ &  $\mathbf{67.6000 \pm 0.3123}$ &  $\mathbf{1.35}$ &  $61.08$ \\
                     & \algoAA{} &  $94.22$ &  $69.2780 \pm 0.3741$ &  $1.39$ &  $67.78$ \\
                     & \algoCN{} &  $93.12$ &  $69.1860 \pm 0.3855$ &  $1.38$ &  $67.48$ \\
                     & \algoJC{} &  $95.36$ &  $69.5460 \pm 0.4242$ &  $1.39$ &  $68.60$ \\
                     & \algoRA{} &  $96.41$ &  $69.4460 \pm 0.3569$ &  $1.39$ &  $69.36$ \\
                     & \algoSR{} &  $88.82$ &  $68.2940 \pm 0.2667$ &  $1.37$ &  $64.83$ \\
\midrule
\multirow{6}{*}{\shortstack[l]{\datachs\\~\\$|V|=327$\\$|F|=7{,}818$}} & \algoPA{} &  $67.05$ &  $56.7160 \pm 0.1669$ &  $1.13$ &  $59.34$ \\
             & \algoAA{} &  $93.69$ &  $54.8560 \pm 0.1876$ &  $1.10$ &  $85.17$ \\
                     & \algoCN{} &  $93.55$ &  $54.8240 \pm 0.1940$ &  $1.10$ &  $85.05$ \\
                     & \algoJC{} &  $93.20$ &  $\mathbf{53.1520 \pm 0.2081}$ &  $\mathbf{1.06}$ &  $87.92$ \\
                     & \algoRA{} &  $93.81$ &  $54.8580 \pm 0.1955$ &  $1.10$ &  $85.28$ \\
                     & \algoSR{} &  $92.50$ &  $56.7240 \pm 0.1166$ &  $1.13$ &  $81.86$ \\
\midrule
\multirow{6}{*}{\shortstack[l]{\datacps\\~\\$|V|=242$\\$|F|=12{,}704$}} & \algoPA{} &  $72.20$ &  $56.1760 \pm 0.2117$ &  $1.12$ &  $64.46$ \\
                     & \algoAA{} &  $86.95$ &  {$52.3760 \pm 0.1852$} &  {$1.05$} &  $82.81$ \\
                     & \algoCN{} &  $86.27$ &  {$52.3560 \pm 0.1929$} &  {$1.05$} &  $82.16$ \\
                     & \algoJC{} &  $88.94$ &  {$\mathbf{49.3960 \pm 0.1983}$} &  {$\mathbf{0.99}$} &  $89.84$ \\
                     & \algoRA{} &  $88.73$ &  {$52.3860 \pm 0.1822$} &  {$1.05$} &  $84.50$ \\
                     & \algoSR{} &  $86.28$ &  {$55.0620 \pm 0.2876$} &  {$1.10$} &  $78.44$ \\
\midrule
\multirow{6}{*}{\shortstack[l]{\datandc\\~\\$|V|=5{,}556$\\$|F|=4{,}525$}} & \algoPA{}${}^{a}$ &  $96.86$ &  $\mathbf{66.8960 \pm 0.1153}$ &  $\mathbf{1.34}$ &  $72.28$ \\
                     & \algoAA{}${}^{a}$ &  $99.13$ &  $96.4960 \pm 0.0806$ &  $1.93$ &  $51.36$ \\
                     & \algoCN{} &  $98.95$ &  $96.1080 \pm 0.0966$ &  $1.92$ &  $51.54$ \\
                     & \algoJC{} &  $98.67$ &  $98.4800 \pm 0.0245$ &  $1.97$ &  $50.09$ \\
                     & \algoRA{}${}^{b}$ &  $99.57$ &  $97.7540 \pm 0.0206$ &  $1.96$ &  $50.80$ \\
                     & \algoSR{}${}^{b}$ &  $98.88$ &  $88.9620 \pm 0.0479$ &  $1.78$ &  $55.55$ \\
\midrule
\multirow{6}{*}{\shortstack[l]{\datatms\\~\\$|V|=1{,}554$\\$|F|=22{,}274$}} & \algoPA{}${}^c$ &  $90.48$ &  {$\mathbf{56.0940 \pm 0.0585}$} &  {$\mathbf{1.12}$} &  $80.79$ \\
                     & \algoAA{}${}^c$ &  $94.88$ &  $64.8120 \pm 0.0611$ &  $1.30$ &  $72.98$ \\
                     & \algoCN{} &  $94.31$ &  $64.6880 \pm 0.0601$ &  $1.29$ &  $73.11$ \\
                     & \algoJC{} &  $89.43$ &  $65.2260 \pm 0.0700$ &  $1.30$ &  $68.79$ \\
                     & \algoRA{}${}^d$ &  $96.04$ &  $64.8480 \pm 0.0631$ &  $1.30$ &  $73.88$ \\
                     & \algoSR{}${}^d$ &  $94.78$ &  $60.0960 \pm 0.0422$ &  $1.20$ &  $78.98$ \\
\bottomrule
\end{tabular}
\end{table*}

\section{Results and Discussion}
Table \ref{tab:relocated_aucs} shows the AUC scores obtained on an original hypergraph $H$ and its relocated versions $H_{rel}$, the adjustment factors $AF$, and the adjusted-AUC scores $AUC_{adj}$ for real-world datasets taken from Benson et al.~\cite{Benson2018} available from this link: \texttt{\url{http://www.cs.cornell.edu/~arb/data/}}.
We perform \textit{five} relocations, and hence report the mean and standard-deviation values for $AUC_{rel}$ (other details regarding reproducibility have been included in the supplementary material).
Following are the LP algorithms used: \textit{Preferential Attachment} (\algoPA)~\cite{Newman2001,jeong2003measuring}, \textit{Adamic Adar} (\algoAA)~\cite{Adamic2003}, \textit{Common Neighbors} (\algoCN)~\cite{Newman2001}, \textit{Jaccard Coefficient} (\algoJC)~\cite{Liben-Nowell2003}, \textit{Resource Allocation} (\algoRA)~\cite{zhou2009predicting}, and \textit{SimRank} (\algoSR)~\cite{jeh2002simrank}.
The following key observations can be made from Table~\ref{tab:relocated_aucs}:
\begin{itemize}
  \item For \datandc{}, wherein we predict drug interactions, the effect of higher-order relations is the highest.
  Without adjustment, all heuristics estimate that they would be able to predict around $96$--$99\%$ of links.
  However, the randomized (relocated) hypergraph also gives a really high score (except for \algoPA{} and \algoSR{}).
  In reality, for most heuristics, the score is only around $50$--$55\%$ as obtained after adjustment (again, \algoPA{} is an exception).
  \item Interestingly the adjustment factors are \textit{proportional to number of hyperedges of higher orders}. That is, higher the number of higher order relations, larger the adjustment factor.
  For instance, from Figure \ref{fig:hyg_sizes}(a), dataset \datacps{} has the least number of higher-sized hyperedges, and also the least of the adjustment factors ($1$--$1.12$).
  On the other hand, dataset \datandc{} has a higher number of higher-sized hyperedges, and hence the largest of adjustment factors ($1.34$--$1.97$).
  \item To better understand the need for an adjustment factor and an adjusted AUC for a hypergraph, we consider two algorithms $A_1$ and $A_2$.
Now, any comparison of these algorithms assumes \textit{similar} AUC scores in case of randomly generated datasets;\footnote{Random in the sense that any useful prediction cannot be made for such datasets. Consider a simple classification problem where both classes $0$ and $1$ come from the same distribution. In such cases, it is known that \textit{\textbf{no}} classifier could be successfully learnt.} this can be considered as a \textit{baseline} for their comparison. 
This seems true for some algorithm pairs in Table~\ref{tab:relocated_aucs}: \textit{e.g.}, \algoAA{} and \algoCN{} share similar $AUC_{rel}$ scores for each dataset.
Contrast this with the pairs marked using superscripts $a$, $b$, $c$, and $d$.
For instance, consider the algorithm pairs (\algoPA, \algoAA) and (\algoRA, \algoSR), which on both the randomly relocated hypergraphs \datandc{} and \datatms{}, \textit{do not perform} equally, in that \algoAA{} and \algoRA{} decently outperform \algoPA{} and \algoSR{} respectively.
Hence, the \textit{baselines are quite different}, and so algorithms \algoPA{} and \algoAA{} are not comparable here.
Same is the case for \algoRA{} and \algoSR{}.
\item A remarkable observation is that the adjusted AUCs of the $a$-, $b$-, $c$-, and $d$-marked algorithm-pairs show a \textit{performance-reversal}, \textit{i.e.}, the AUC order reverses for algorithm-pairs (\algoPA, \algoAA) and (\algoRA, \algoSR) when adjusted.
More specifically, for both datasets \datandc{} and \datatms{}, we have $AUC(\algoAA) > AUC(\algoPA)$, but $AUC_{adj}(\algoAA) < AUC_{adj}(\algoPA)$ (similarly, $AUC(\algoRA) > AUC(\algoSR)$, but $AUC_{adj}(\algoRA) < AUC_{adj}(\algoSR)$), reversing the performance rating of the algorithms.
This is indeed the situation which occurs in presence of higher-order relations, and hence correction is required for proper evaluation.
As link prediction is a 2-class problem, the appropriate baseline is indeed $0.5$, and one should normalize the scores accordingly.
This is achieved by the adjustment factor.
\end{itemize}
\section{Conclusion And Future Work}
\label{sec:5}
To summarize, we have proved in this article that \textit{higher-order relations skew link prediction} in simple graphs.
This is achieved by proposing a simple model for hypergraphs. Using this model, we show that the LP algorithms such as CN/AA \textit{do not generalize well} in presence of higher-order relations.
Moreover, we prove that these algorithms also tend to \textit{overestimate their ability to predict links}.
We correct this by proposing a \textit{new evaluation approach} by computing an \textit{adjustment factor}.

The main insight obtained in this article -- higher-order relations skew link prediction on graphs -- has much bigger consequences than can be discussed within the scope of this article.
Firstly, the broader question remains: Why do higher-order relations even affect link prediction?
And how can one even correct it in general?
There is no simple answer to this question at this point.
We believe that the fundamental reason why higher-order relations have this effect is that there is no unambiguous way to model higher-order relations in simple graphs.
Any approach to convert hypergraphs into simple graphs either loses information or adds bias or both.
We also hypothesize that the effect of higher-order relations is not limited to link prediction and they affect several other problems pertaining to networks as well.
These constitute the main directions for future research. 


\section*{Broader Impact (As required by NeurIPS)}

To our understanding, the societal impact of this current research is not explicit, but is implicit.
Understanding and modelling higher-order relations could potentially allow for identification/correction of biases in network-based tasks.
For instance consortiums (which are modelled as higher-order relations) may knowingly or unknowingly induce bias into the model.
A deeper understanding of higher-order relations would help identify/correct these effects.

Link prediction is widely used in important applications such as author-ranking, recommendation systems, social network analysis, \textit{etc}.
However, the impact of the introduction of a bias in the working of link prediction algorithms might not hurt much.
On the other hand, in more serious applications such as protein-protein interaction, protein-disease interaction, disease-disease comorbidity, drug-drug reaction, \textit{etc}., the effects could be more concerning.
Most of these networks occur mainly as hypergraphs, since the corresponding relations are of a higher-order nature (\textit{e.g.}, three drugs could be harmful if taken together, but pairs of the same might not).
While link prediction refers to only pairs of such entities, the effect of the underlying higher-order relations could not be ignored.

For example, in an emergency scenario, a link prediction heuristic might predict biased protein-disease interactions due to the presence of an otherwise-ignored higher-order structure among the proteins and diseases.
\textit{If not for analyses like the one done in the present article, such seemingly biased predictions might turn fatal}.
Even if for some particular datasets, the heuristics do not introduce any bias despite an underlying hypergraph structure being present, there is no harm in performing a parallel analysis of the effect they could have on different versions of the data.

In summary, relational learning is a \textit{wide area of research}, and touches many important real-world problems.
Moreover, almost all research on relations happen with the \textit{assumption of pairwise connections} between entities, which is seldom true (since most of them actually form from higher-order relations).
Both these factors highlight the impact our present work could have on most real-world applications.

\bibliographystyle{plain}
\bibliography{ms}

\appendix

\section{Proofs}
\label{app:proofs}

\subsection{Proof of Proposition \ref{prop:1}}
\label{app:proof:prop}

  Given $i, j \in V$, and $f \in \overline{F}$, define $S_k(i, j, f) := 1$ if $f$ has $\{i,j\}$ as a subset of its vertices; otherwise, $S_k(i, j, f) = 0$.
  We hence have $S_k(i,j) = \sum_{f \in \overline F} S_k(i,j,f)$. Note that the event $(i \not\sim j)$ holds if and only when all the events $(i \not\sim j, S_k(i,j,f)=1)$ for all $k$ hold and each of these events are mutually independent, a fact known from the model.
  Hence, 
  \begin{equation}
      \label{eq:prob-generalization}
      P(i \not\sim j) = \Pi_{f}  P(i \not\sim j, S_k(i,j,f)=1).
  \end{equation}
  Now, since $S_k(i,j,f)$ is either $1$ or $0$, and depends on $U$ and $r$, it is independent of the event $(i \not\sim j)$. Hence we have, 
  \begin{eqnarray*}
      P(i \not\sim j) &=& \Pi_{f}  P(i \not\sim j / S_k(i,j,f)=1) \cdot S_k(i,j,f) \\
       &=& \Pi_{k} (1 - \phi_k)^{S_k(i,j)}.
  \end{eqnarray*}
  
\subsection{Proof of Theorem \ref{thm:1}}
\label{app:proof:1}
\begin{proof}
Let $\{i,j\}$ denote arbitrary but fixed pair of points such that there exists a unique hyperedge $h \supset \{i,j\}$. We prove that in this case, 
\begin{equation}
    P\left(Z_{i \sim j} > Z_{i \not\sim j} / Z_{i \sim j} \neq Z_{i \not\sim j}\right) > 0.5
\end{equation}
The proof for the generic case is similar. 

\textbf{Claim 1 :} We first show that, 
\begin{equation}
    P(Z_{i \sim j} = 0) < P(Z_{i \not\sim j} = 0)
\end{equation}
and for all $k \geq 1$,
\begin{equation}
    P(Z_{i \sim j} > k) > P(Z_{i \not\sim j} > k)
\end{equation}

Recall that $\overline{F}$ denotes the set of all possible hyperedges. Now, an event $E$ is essentially selecting the subset of $\overline{F}$. Let $E_{i \not\sim j}$ denote an event where $i \not\sim j$. Then, one can construct the event $E_{i \sim j}$ by ensuring that one picks $h$. So, if $P(E_{i \not\sim j}) = p$, then, $P(E_{i \sim j}) = p*\phi_{\lvert h \rvert}/(1 - \phi_{\lvert h \rvert})$.
Now, 
\begin{equation}
\begin{split}
    P(E_{i \not\sim j}/ i\not\sim j) &= P(E_{i \not\sim j})/P( i\not\sim j) = p/(1 - \phi_{\lvert h \rvert}) \\ 
    &= p\phi_{\lvert h \rvert}/(\phi_{\lvert h \rvert}(1 - \phi_{\lvert h \rvert})) = P(E_{i \sim j})/P(i \sim j) \\ 
    &=  P(E_{i \sim j}/i \sim j)
\end{split}
\end{equation}
Moreover we have that $CN(E_{i\sim j})$ (Common neighbors of $\{i,j\}$ when $i\sim j$) is greater than or equal to $CN(E_{i\not\sim j})$. Now, consider $P(Z_{i \sim j} = 0)$. Since, if $i \sim j $, then $h$ is selected and $\lvert h \rvert > 2$, we must have that $P(Z_{i \sim j} = 0) = 0$ that is there exists at least one common neighbor for $\{i,j\}$ when $i \sim j$. Clearly, it is possible that $Z_{i \not\sim j} = 0$ and hence 
\begin{equation}
    P(Z_{i \sim j} = 0) < P(Z_{i \not\sim j} = 0)
\end{equation}
Now, consider the case when $k \geq 1$, 
\begin{eqnarray}
 P(Z_{i \not\sim j} > k) &=& \sum_{E_{i \not\sim j}} I\{CN(E_{i \not\sim j})  > k\} P(E_{i \not\sim j}/ i\not\sim j) \\
 & < & \sum_{E_{i \sim j}} I\{CN(E_{i \sim j}) > k\} P(E_{i \sim j}/ i\sim j) \\
 & = & P(Z_{i \sim j} > k)
\end{eqnarray}
Since, as shown earlier, for each $E_{i \not\sim j}$, one can construct a corresponding $E_{i \sim j}$ such that CN($E_{i \sim j}$) is greater than or equal to CN($E_{i \not\sim j}$). The strong inequality holds since one can construct atleast one case such that  CN($E_{i \sim j}$) is strictly greater than CN($E_{i \not\sim j}$). (check this carefully)

Now, let $p_k = P(Z_{i \sim j} = k)$ and $p_{k}^\prime = P(Z_{i \not\sim j} = k)$
\begin{eqnarray}
AUC & = & P\left(Z_{i \sim j} > Z_{i \not\sim j}/Z_{i \sim j} \neq Z_{i \not\sim j}\right) \\
& = &\sum_k \frac{P\left(Z_{i \sim j} > Z_{i \not\sim j}\right)}{P\left(Z_{i \sim j} \neq Z_{i \not\sim j}\right)}\\
&=& \sum_{k}{\frac{P\left(Z_{i \sim j} > k, Z_{i \not\sim j} = k\right)}{\sum_{k^\prime}P\left(Z_{i \sim j} \neq {k^\prime}, Z_{i \not\sim j} = {k^\prime}\right)}}\\
&=& \sum_k \frac{P\left(Z_{i \sim j} > k \right) P\left( Z_{i \not\sim j} = k\right)}{\sum_{k^\prime} P\left(Z_{i \sim j} \neq {k^\prime}\right) P\left( Z_{i \not\sim j} = {k^\prime}\right)}\\
& > & \sum_k \frac{P\left(Z_{i \not\sim j} > k \right) P\left( Z_{i \not\sim j} = k\right)}{\sum_{k^\prime} P\left(Z_{i \sim j} \neq {k^\prime}\right) P\left( Z_{i \not\sim j} = {k^\prime}\right)}\\
& = &  \sum_k \frac{P\left(Z_{i \not\sim j} < k \right) P\left( Z_{i \not\sim j} = k\right)}{\sum_{k^\prime} P\left(Z_{i \sim j} \neq {k^\prime}\right) P\left( Z_{i \not\sim j} = {k^\prime}\right)}\\
\end{eqnarray}

Now, for every event $E_{i \sim j}$ one can construct an event $E_{i \not\sim j}$ by simply removing $h$. Moreover, if $CN(E_{i \sim j}) < k$ then, it implies that $CN(E_{i \not\sim j}) < k$. Hence, $P\left(Z_{i \not\sim j} < k \right) > P\left(Z_{i \sim j} < k \right)$. So, we have  

\begin{eqnarray}
AUC & > &  \sum_k \frac{P\left(Z_{i \not\sim j} < k \right) P\left( Z_{i \not\sim j} = k\right)}{\sum_{k^\prime} P\left(Z_{i \sim j} \neq {k^\prime}\right) P\left( Z_{i \not\sim j} = {k^\prime}\right)}\\
& > &  \sum_k \frac{P\left(Z_{i \sim j} < k \right) P\left( Z_{i \not\sim j} = k\right)}{\sum_{k^\prime} P\left(Z_{i \sim j} \neq {k^\prime}\right) P\left( Z_{i \not\sim j} = {k^\prime}\right)}\\
& = &  P\left(Z_{i \sim j} < Z_{i \not\sim j}/Z_{i \sim j} \neq Z_{i \not\sim j}\right)
\end{eqnarray}

Hence we have that,
\begin{equation}
    AUC = P\left(Z_{i \sim j} > Z_{i \not\sim j}/Z_{i \sim j} \neq Z_{i \not\sim j}\right) > 0.5
\end{equation}
as, 
\begin{equation}
    P\left(Z_{i \sim j} > Z_{i \not\sim j}/Z_{i \sim j} \neq Z_{i \not\sim j}\right) + P\left(Z_{i \sim j} < Z_{i \not\sim j}/Z_{i \sim j} \neq Z_{i \not\sim j}\right) = 1
\end{equation}
Hence, proved.
\end{proof}

\subsection{Proof of Theorem \ref{thm:2}}
\label{app:proof:2}
\begin{proof}
Given $n \in \mathbb{N}$ and $p \in \mathbb{R}$, we have $V = \{1, 2, \hdots, n\}$ and $P(i\sim j) = p$. 
Suppose there is a link predictor $\pi:\powerset_2(V) \rightarrow \mathbb{R}$. 
If $Z_\sim := (\pi(\{i, j\})\mid i \sim j)$ and $Z_\nsim := (\pi(\{i, j\})\mid i\nsim j)$, we have $AUC = P(Z_\sim > Z_\nsim\mid Z_\sim \neq Z_\nsim)$. 
Now, whatever be the logic the value of $\pi$ depends upon, it would not differentiate between links and non-links since the ``environment'' for link prediction was formed at random. 
In other words, $\pi(\{i, j\})$ would follow the same distribution for both links and non-links, giving us $P(Z_\sim = k) = P(Z_\nsim = k)~\forall k$ (assuming them to be discrete random variables; a similar argument holds for a continuous one as well). 
So, we have $P(Z_\sim > Z_\nsim \mid Z_\sim \neq Z_\nsim) = P(X > Y\mid X\neq Y) = 0.5$ (where $X$ and $Y$ are two random variables from the same distribution).
\end{proof}

\end{document}